# FCOSR: A Simple Anchor-free Rotated Detector for Aerial Object Detection

Zhonghua Li, Biao Hou, Zitong Wu, Licheng Jiao, Bo Ren, Chen Yang
School of Artificial Intelligence, Xidian University, Xi'an, China
zhli_1991@stu.xidian.edu.cn, avcodec@163.com, wuzitong@stu.xidian.edu.cn,
lchjiao@mail.xidian.edu.cn, asdwer2046@126.com, 18792675001@163.com

## Abstract

*Existing anchor-base oriented object detection methods have achieved amazing results, but these methods require some manual preset boxes, which introduces additional hyperparameters and calculations. The existing anchor-free methods usually have complex architectures and are not easy to deploy. Our goal is to propose an algorithm which is simple and easy-to-deploy for aerial image detection. In this paper, we present a one-stage anchor-free rotated object detector (FCOSR) based on FCOS, which can be deployed on most platforms. The FCOSR has a simple architecture consisting of only convolution layers. Our work focuses on the label assignment strategy for the training phase. We use ellipse center sampling method to define a suitable sampling region for oriented bounding box (OBB). The fuzzy sample assignment strategy provides reasonable labels for overlapping objects. To solve the insufficient sampling problem, a multi-level sampling module is designed. These strategies allocate more appropriate labels to training samples. Our algorithm achieves 79.25, 75.41, and 90.15 mAP on DOTA1.0, DOTA1.5, and HRSC2016 datasets, respectively. FCOSR demonstrates superior performance to other methods in single-scale evaluation. We convert a lightweight FCOSR model to TensorRT format, which achieves 73.93 mAP on DOTA1.0 at a speed of 10.68 FPS on Jetson Xavier NX with single scale. The code is available at:* <https://github.com/lzh420202/FCOSR>

## 1. Introduction

The object detection task usually uses horizontal bounding box (HBB) to circle the target and give its category. In recent years, many excellent HBB framework algorithms have been proposed, including YOLO series[1-3], R-CNN series[4-7], RetinaNet[8], FCOS[9], and CenterNet[10], etc. These methods have achieved amazing results in object detection tasks. There are many challenges in single-image aerial object detection task, such as arbitrary orientation, dense objects, and wide range of resolution. These above problems make the HBB algorithm difficult to detect aerial objects effectively. Therefore, the aerial object detection task converts HBB into an oriented bounding box (OBB) by adding a rotation angle. At present, oriented object detector is generally modified from HBB algorithms, which can be divided into two types: anchor-base methods[11-23], and anchor-free methods[24-33]. The anchor-base methods usually require manual preset boxes, which not only introduces additional hyperparameters and calculations, but also directly affects the performance of the model. The anchor-free methods remove the preset box and reduce the prior information, which makes it more adaptable than the anchor-base methods. In this paper, we propose a one-stage anchor-free rotated object detector (FCOSR) based on FCOS[9] and 2-dimensional (2D) gaussian distribution. Our method directly predicts the center point, width, height, and angle of the object. The main work of this paper focuses on training stage. Benefitting from the redesigned label assignment strategy, our method can predict OBB of the target directly and accurately with few modifications to the FCOS architecture. Compared with the refined two-step methods, our method is not only simpler, but also has only convolutional layers, so it is easier to deploy to most platforms. A series of experiments on DOTA[34] and HRSC2016[35] datasets verify the effectiveness of our method. Our contributions are as: **(1)** We propose a one-stage anchor-free aerial oriented object detector, which is simple, fast and easy in deployment. **(2)** We design a set of label assignment strategies based on 2D gaussian distribution and aerial image characteristics. These strategies assign more appropriate labels to training samples. **(3)** Our method achieves **79.25**, **75.41**, and **90.15** mAP on DOTA1.0, DOTA1.5, and HRSC2016 datasets, respectively. Compared with other anchor-free methods, FCOSR achieves state-of-the-arts. Compared with other anchor-base methods, FCOSR surpasses many two-stage methods in terms of single scale. Our model greatly reduces the gap between anchor-free and anchor-base methods. In terms of speed and accuracy, FCOSR presents its excellent performance, and it surpasses the current mainstream models. **(4)** We convert a lightweight FCOSR to TensorRT format and successfully migrate it to Jetson Xavier NX. The TensorRT model achieves **73.93** mAP with **10.68** FPS on DOTA1.0 test set.

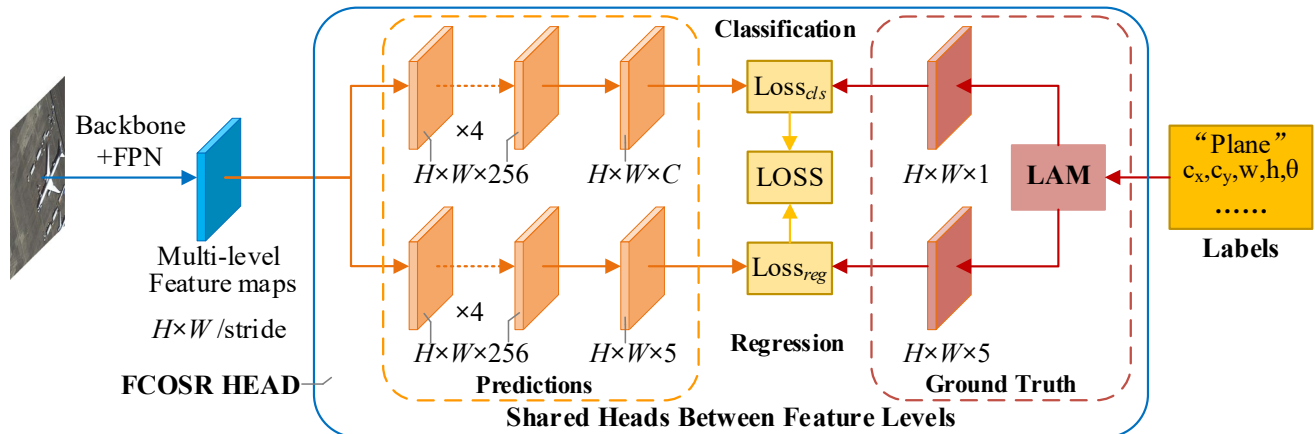

Figure 1. **FCOSR architecture.** The output of backbone with feature pyramid network (FPN)[36] are multi-level feature maps, including $P_3$-$P_7$. The head is shared with all multi-level feature maps. The predictions on the left of the head is the inference part, the other components are only effective during the training stage. LAM means label assignment module, which allocates labels to each feature maps. H and W are height and width of feature map. Stride is the down-sampling ratio for multi-level feature maps. C represents the number of categories, and regression branch directly predicts the center point, width, height and angle of the target.

## 2. Related Works

The current mainstream oriented object detection algorithms can be divided into two types, one is the anchor-base methods[11-23], the other is the anchor-free methods[24-33].

### 2.1. Anchor-base methods

The anchor-base methods need to manually preset a series of standard boxes (anchor) for boundary regression and refinement. Early methods used anchors with multiple angles and multiple aspect ratios to detect oriented objects[11,12]. However, the increase of the preset angles leads to a rapid increase of anchors and calculations, which makes the model difficult to train. As a two-stage method, ROI transformer[13] converts the horizontal proposal into OBB format through the RROI learning module, then extracts the features in the rotation proposal for subsequent classification and regression. This method replaces the preset angles by giving the angle value through the network, which greatly reduces anchors and calculations. For a long time, many ROI-transformer-base methods have appeared and achieved good results. ReDet[14] introduces rotation invariant convolution (e2cnn)[37] into the whole model and extracts rotation invariant features by using RiROI alignment. Oriented R-CNN[15] replaces RROI learning module in ROI-transformer[13] with a lighter and simpler oriented region proposal network (orientation RPN). $R^3$Det[18] is a refined one-stage oriented object detection method, which obtains OBB result by fine tuning the anchor in HBB format through feature innovation module (FRM) module. $S^2$ANet[19] is composed of feature alignment module (FAM) and oriented detection module (ODM). FAM generates high-quality OBB anchor. ODM adopts active rotating filters to produce orientation-sensitive and orientation-invariant features to alleviate the inconsistency between classification score and localization accuracy. CSL[20] converts angle prediction into a classification task to solve the problem of discontinuous rotation angles. DCL[21] uses dense coding on the basis of CSL[20] to improve training speed. It also uses angle distance and aspect ratio sensitive weighting to improve accuracy.

### 2.2. Anchor-free methods

Current anchor-free methods are mostly one-stage architecture. IENet[24] develops a branch interactive module with self-attention mechanism, which can fuse features from classification and regression branches. The anchor-free methods directly predict the bounding box of the target, which makes the loss design in the regression task have certain limitations. GWD[22], KLD[23], and ProbIoU[25] use the distance metric between two 2D gaussian distributions to represent loss, which provides a new regression loss scheme for anchor-free methods. PIoU[26] designs an IoU loss function for OBB based on pixel statistics. BBAVectors[27] and PolarDet[28] define OBB with bbav vector and polar coordinates respectively. CenterRot[29] uses deformable convolution (DCN)[38] to fuse multi-scale features. AROA[30] leverages attention mechanisms to refine the performance of remote sensing object detection in a one-stage anchor-free network framework.

## 3. FCOSR

As shown in Figure 1. our method takes the FCOS[9] architecture as baseline. The network directly predicts center point (include *x* and *y*), width, height, and rotated angle of target (OpenCV format). We determine the convergence target of the feature map through the label assignment module. Our algorithm does not introduce additional components into the architecture. It removes centerness branch[9], which makes network simpler and easier in deployment. The work of this paper focuses on the label assignment in training stage.

### 3.1. Network outputs

Network output contains a C-dimensional vector from classification branch and a 5-dimensional (5D) vector from regression branch. Unlike FCOS[9], we want each components of regression output to have different ranges. The offset can be negative, width and height must be positive, and angle must be limited to 0-90. These simple processes are defined by Eq. 1.

$$\begin{aligned} offset_{xy} &= Reg_{xy} \cdot k \cdot s \\ wh &= (\text{Elu}(Reg_{wh} \cdot k) + 1) \cdot s \\ \theta &= \text{Mod}(Reg_\theta, \pi/2) \end{aligned} \quad (1)$$

$Reg_{xy}$, $Reg_{wh}$, and $Reg_\theta$ indicate the direct output from last layer of regression branch. $k$ is a learnable adjustment factor and $s$ is the down-sampling ratio (stride) for multi-level feature maps. Elu[39] is the improvement of ReLu.

Through the calculation of above equation, the output will be converted into a new 5D vector ($offset_x$, $offset_y$, $w$, $h$, $angle$). The sampling point coordinates plus offsets to obtain target OBBs.

### 3.2. Ellipse center sampling

Center sampling is a strategy to concentrate sampling points close to the center of target, which helps to reduce low-quality detection and improves model performance. This strategy is adopted in FCOS[9], YOLOX[3], and other networks, which stably improves the accuracy. However, there are two problems when directly migrating horizontal center sampling strategy to oriented object detection. First, the horizontal center sampling area is usually a 3×3 or 5×5 square[3,9], so the angle of OBB will affect the sampling range. Second, short edge further reduces the number of sampling points for large aspect ratio targets. The most intuitive center sampling should be a circular area within a certain range at the center of target, but the short edge limits the range of center sampling. In order to reduce these negative influences, we propose an elliptical center sampling method (ECS) based on 2D gaussian distribution. We use OBB ($cx$, $cy$, $w$, $h$, $\theta$) parameters to define a 2D gaussian distribution[25]:

$$\begin{aligned} \Sigma &= R_\theta \cdot \Sigma_0 \cdot R_\theta^T \\ \mu &= (cx, cy) \\ R_\theta &= \begin{bmatrix} \cos\theta & -\sin\theta \\ \sin\theta & \cos\theta \end{bmatrix}, \Sigma_0 = \frac{1}{12}\begin{bmatrix} w^2 & 0 \\ 0 & h^2 \end{bmatrix} \end{aligned} \quad (2)$$

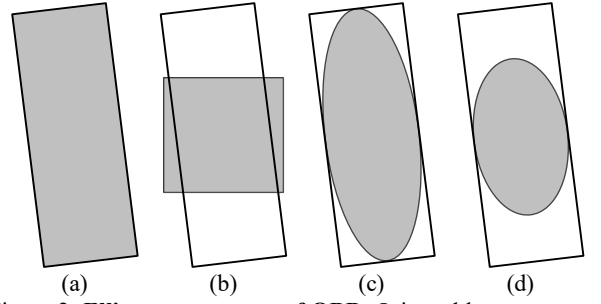

Figure 2. **Ellipse center area of OBB.** Oriented box represents OBB of the target, and the shadow area represents the sampling region. (a) general sampling region, (b) horizontal center sampling region, (c) original elliptical region, (d) shrink elliptical region.

$\Sigma$ is covariance matrix, $\Sigma_0$ is the covariance matrix when angle equal to 0, $\mu$ is mean value, and $R_\theta$ is the rotation transformation matrix.

The contour of the probability density function of 2D gaussian distribution is an elliptic curve. Eq. 3 represents the probability density of 2D gaussian distribution in general case.

$$f(X) = \frac{1}{2\pi|\Sigma|^{1/2}}\exp(-\frac{1}{2}(X-\mu)^T\Sigma^{-1}(X-\mu)) \quad (3)$$

$X$ indicates coordinates (2D vector). We remove the normalization term from $f(X)$ and get $g(X)$.

$$g(X) = \exp(-\frac{1}{2}(X-\mu)^T\Sigma^{-1}(X-\mu)) \quad (4)$$

$g(X) \in (0,1]$, the elliptic contour of 2D gaussian distribution can be expressed as $g(X) = C$.

When $C = C_0 = \exp(-1.5)$, the elliptical contour line is just inscribed in OBB. The range of elliptic curve will expand with the decrease of $C$, which means effective range of $C$ is $[C_0, 1]$. Considering that there are many small objects in aerial images, we set $C$ as 0.23 to prevent insufficient sampling caused by small sampling area. The center sampling area of the target can be determined by $g(X) \geq C$. If $g(X)$ is greater than $C$, the point $X$ is in sampling area. The elliptical area defined by the target with large aspect ratio has a slender shape, which makes the part in long axis direction is far away from the center area. In order to solve this problem, we shrink the ellipse sampling region by modifying the gaussian distribution. Eq. 5 defines new original covariance matrix.

$$\Sigma_0 = \frac{\min(w,h)}{12}\begin{bmatrix} w & 0 \\ 0 & h \end{bmatrix} \quad (5)$$

The length of ellipse major axis shrinks to $\sqrt{wh}$ and minor axis remain unchanged. Figure 2 shows the ellipse center area of OBB. Compared with the horizontal center sampling, the ellipse center sampling is more suitable for OBB, and sampling area of large aspect ratio target is more concentrated by shrinking long axis.

## 3.3. Fuzzy sample label assignment

FCOS[9] reduces fuzzy (ambiguous) samples by allocating different scales targets to feature maps with different strides. For targets with similar scales, FCOS[9] assigns smaller target label to ambiguous sampling points. Obviously, this fuzzy sample label assignment method based on the minimum area principle is difficult to deal with complex scenes, such as aerial scenes. We design a fuzzy sample label assignment method (FLA) to assign ambiguous sample labels based on 2D gaussian distribution. The gaussian distribution presents a bell shape, and the response is the largest in its center. The response becomes smaller as the sampling point moves away from the center of the distribution. We approximately take the 2D gaussian distribution as distance measure between sampling point and target center. The center distance $J(X)$ is defined by Eq. 6.

$$J(X) = \sqrt{wh} \cdot f(X) \qquad (6)$$

$f(X)$ is probability density of 2D gaussian distribution, which is defined in Eq. 3.

For any target, we calculate the $J(X)$ value at each sampling point. A larger value of $J(X)$ means that $X$ is closer to the target. When a sampling point is included by multiple targets at the same time, we assign the label with the largest $J(X)$ target to sampling point. A simple fuzzy sample label assignment diagram is shown in Figure 3.

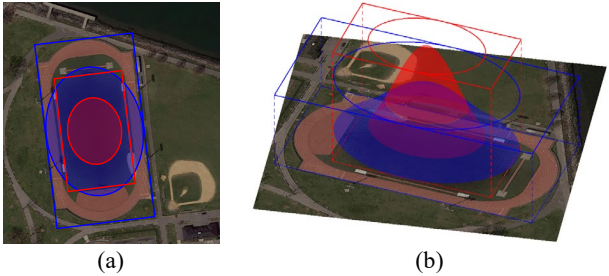

(a)          (b)

Figure 3. **A Fuzzy sample label assignment demo.** (a) is a 2D label assignment area diagram, and (b) is a 3D visualization effect diagram of $J(X)$ of two targets. The red OBB and area represent the court target, and the blue represents the ground track field. After $J(X)$ calculation, smaller areas in the red ellipse will be allocated to the court, and other blue areas will be allocated to the ground track field.

## 3.4. Multi-level sampling

The sampling range of large aspect ratio targets is mainly affected by the short edge. As shown in Figure 4, when the stride of feature map is greater than the length of short edge, target may be too narrow to be effectively sampled. Therefore, in the view of possibility of insufficient sampling, we add a simple supplementary sampling scheme, which determines whether allocate labels in the lower-level feature map by comparing short edge with stride. We assign labels to feature maps that satisfy the following two conditions. First, the ratio between the short edge of the target and the stride of the feature map is less than 2. Second, the long edge of minimum bounding rectangle of the target is larger than the acceptance range of feature map. Multi-level sampling strategy (MLS) allows us to add some targets that cannot be effectively sampled to the lower-level feature map. The lower-level feature map represents denser sampling points, which alleviates the insufficient sampling problem.

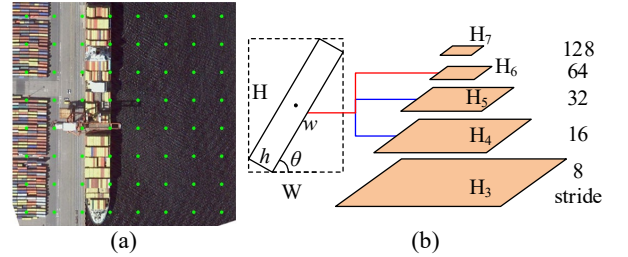

(a)          (b)

Figure 4. **Multi-level sampling**. (a) insufficient sampling, green points in diagram are sampling points. The ship is so narrow that there are no sampling points inside it. (b) is a multi-level sampling demo. The red line indicates that the target follows FCOS guidelines assigned to $H_6$ while it is too narrow to sample effectively. The blue line indicates that the target is assigned to lower level of features according to the multi-level guideline. This represents the target sampling at three different scales to handle the insufficient sampling problem.

## 3.5. Target Loss

The loss of FCOSR consists of classification loss and regression loss. Quality focal loss (QFL)[40] is used for classification loss, which is mainly remove the centerness branch from original FCOS[9] algorithm. The regression uses the ProbIoU loss[25].

QFL[40] is a part of general focal loss (GFL)[40]. It unifies training and testing process by replacing the one hot label with IOU value between prediction and ground truth. QFL[40] suppresses low-quality detection results and also improves the performance of the model. Eq. 7 gives the definition of QFL[40].

$$QFL(\sigma) = -|y - \sigma|^{\beta}\left((1-y)\log(1-\sigma) + y\log(\sigma)\right) \qquad (7)$$

$y$ represents the replaced IOU, parameter $\beta$ (using the recommend value 2) controls the down-weighting rate smoothly.

ProbIoU loss[25] is a kind of IoU loss specially designed for oriented object. It mainly represents the IOU between OBBs through the distance between 2D gaussian distributions, which is similar to GWD[22] and KLD[23]. The overall loss can be defined by Eq. 8.

$$Loss = \frac{1}{N_{pos}}\sum_{z} QFL + \frac{1}{\sum_{z}\mathbf{1}_{\{c_z^* > 0\}} IoU} \cdot \sum_{z}\mathbf{1}_{\{c_z^* > 0\}} IoU \cdot Loss_{ProbIoU} \qquad (8)$$

$N_{pos}$ represents the number of positive samples. The summation is calculated over all locations $z$ on the multi-level feature maps. $\mathbf{1}_{\{c_z^* > 0\}}$ is the indicator function, being 1 if $c_z^* > 0$ and 0 otherwise.

## 4. Experiments

### 4.1. Datasets

We evaluate our method on DOTA1.0, DOTA1.5, and HRSC2016 datasets.

**DOTA**[34] is a large-scale dataset for aerial object detection. The data is collected from different sensors and platforms. DOTA1.0 contains 2806 large aerial images with the size ranges from 800×800 to 4000×4000 and 188,282 instances among 15 common categories: Plane (PL), Baseball diamond (BD), Bridge (BR), Ground track field (GTF), Small vehicle (SV), Large vehicle (LV), Ship (SH), Tennis court (TC), Basketball court (BC), Storage tank (ST), Soccer-ball field (SBF), Roundabout (RA), Harbor (HA), Swimming pool (SP), and Helicopter (HC). DOTA1.5 adds the Container Crane (CC) class and instances smaller than 10 pixels on the basis of version 1.0. DOTA1.5 contains 402,089 instances. Compared with DOTA1.0, DOTA1.5 is more challenging but stable during training.

In our paper, we use both train and validation sets for training and use test set for testing. All images are cropped into 1024×1024 patches with the gap of 512, and the multi-scale argument of DOTA1.0 are {0.5, 1.0}, while those of DOTA1.5 are {0.5, 1.0, 1.5}. We also apply random flipping and random rotation argument method during training.

**HRSC2016**[35] is a challenging ship detection dataset with OBB annotations, which contains 1061 aerial images with the size ranges from 300×300 to 1500×900. It includes 436, 181, and 444 images in the train, validation, and test set, respectively. We use both train and validation set for training and the test set for testing. All images are resized to 800×800 without changing the aspect ratio. Random flipping and random rotation are applied during training.

### 4.2. Implement Details

We adopt ResNext50[42] with FPN[36] as the backbone for FCOSR. We train the model in 36 epochs for DOTA and 40k iterations for HRSC2016. We use SGD optimizer to train model of DOTA with an initial learning rate (LR) of 0.01, and LR is divided by 10 at {24, 33} epoch. The initial LR of HRSC2016 model is set to 0.001, and step is {30k, 36k} iterations. The momentum and weight decay are 0.9 and 0.0001, respectively. We use Nvidia DGX Station (4 V100 GPUs@32G) with a total batch size 16 for training, and use single 2080Ti GPU for test. We adopt Jetson Xavier NX with TensorRT as embedded deployment platforms. The NMS threshold is set to 0.1 when merging image patches and confidence threshold is set to 0.1 during testing. Inspired by rotation-equivariant CNNs[14,37], we adopt a new rotation augmentation method, which uses 2 step rotation to generate random augmentation data. First, we rotate image randomly in 0, 90, 180 and 270 degrees with equal probability. Second, we rotate image randomly in 30, 60 degrees with 50% probability. Our implement is based on mmdetection[41].

### 4.3. Ablation Studies

We perform a series of experiments on the DOTA1.0 test set to evaluate the effectiveness of the proposed method. We use ResNext50[42] as backbone and call this model FCOSR-M. Other models in FCOSR series can be found in the next section. We training and testing the model with single scale.

As shown in Table 1, the mAP at baseline for FCOS-M is 70.4, which increases by 4.03 with the addition of rotation augmentation. When QFL[40] is used instead of focal loss, the detection result of the model gains 0.91 mAP. Then we try to add the ECS and FLA modules, which increases the detection result to 76.80 mAP. Finally, we add the MLS module and achieve 77.15 mAP on DOTA1.0 with single scale. Through the use of multiple modules, FCOSR-M achieves a very significant performance improvement. These modules do not have any additional calculations during inferencing, which makes FCOSR a very simple, fast, and easy deployment OBB detector.

| Methods | Rot | QFL | ECS | FLA | MLS | mAP (%) |
|---|---|---|---|---|---|---|
| FCOSR-M single scale | | | | | | 70.40 |
| | ✓ | | | | | 74.43 (+4.03) |
| | ✓ | ✓ | | | | 75.34 (+0.91) |
| | ✓ | ✓ | ✓ | ✓ | | 76.80 (+1.46) |
| | ✓ | ✓ | ✓ | ✓ | ✓ | 77.15 (+0.35) |

Table 1. **The result of ablation experiments for FCOSR-M.** ✓ means the module is used. Rot means 2 step rotation augmentation. QFL means quality focal loss. when not using QFL, we use focal loss instead. ECS means ellipse center sampling. FLA means fuzzy sample label assignment module. MLS means multi-level sampling.

### 4.4. More Backbones

We use a variety of other backbones to replace ResNext50[42] to reconstruct the FCOSR model. We adopt Mobilenetv2[43] as backbone, and named FCOSR-S. We also test FCOSR on ResNext101[42] with 64 groups and 4 width, and named FCOSR-L. Model size, FLOPs, FPS, and mAP on DOTA is shown in Table 2.

In order to deploy FCOSR on the embedded platform, we perform lightweight processing on the model. We adjust the output stage of backbone on the basis of FCOSR-

| Methods | backbone | Parameters | GFLOPs | FPS | mAP (%) |
|---|---|---|---|---|---|
| FCOSR-S | Mobile v2 | 7.32M | 101.42 | 23.7/7.1 | 74.05 |
| FCOSR-M | X50-32×4 | 31.4M | 210.01 | 14.6/4.1 | 77.15 |
| FCOSR-L | X101-64×4 | 89.64M | 445.75 | 7.9/2.1 | 77.39 |

Table 2. **FCOSR series model size, FLOPs, FPS and mAP comparison.** X means ResNext[42]. Mobile v2 means Mobilenet v2[43]. FLOPs and FPS is result which test with 1024×1024 image size and single 2080Ti device, and second FPS represents the speed of the rotation test mode. mAP is the result of DOTA1.0 with single-scale evaluation.

| Methods | Parameters | Model size | GFLOPs | FPS | mAP (%) |
|---|---|---|---|---|---|
| FCOSR-lite | 6.9M | 51.63MB | 101.25 | 7.64 | 74.30 |
| FCOSR-tiny | 3.52M | 23.2MB | 35.89 | 10.68 | 73.93 |

Table 3. **Lightweight FCOSR test result on Jetson Xavier NX.** Model size is TensorRT engine file size (.trt). FPS is the average speed of 2000 inferences with 1024×1024 image size. The mAP is the result on DOTA1.0 single-scale test set.

S, and replace the extra convolutional layer of FPN with pooling layer. We call it FCOSR-lite. On this basis, we further adjust the feature channel of the head from 256 to 128, and name it FCOSR-tiny. The above two models are converted to TensorRT 16-bit format and tested on Jetson Xavier NX. The result is shown in Table 3. The lightweight FCOSR achieves perfect balance between speed and accuracy on Jetson Xavier NX. The lightest tiny model achieves **73.93** mAP at **10.68** FPS. This is a successful attempt to deploy a high-performance oriented object detector on edge computing devices.

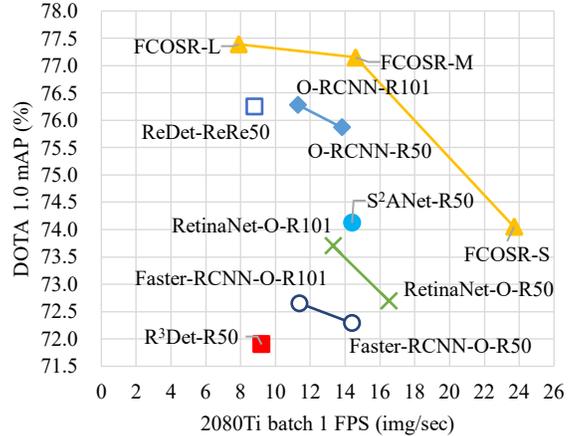

Figure 5. **Speed vs accuracy on DOTA 1.0 test set.** We test ReDet[14], S$^2$A-Net[19] and R$^3$Det[18] on single 2080ti device based on their source code. Faster-RCNN-O[6], RetinaNet-O[8], and Oriented RCNN(O-RCNN)[15] test results are from the OBBDetection. (https://github.com/jbwang1997/OBBDetection)

### 4.5. Speed VS Accuracy

We test the inference speed of FCOSR series models and other open source mainstream models, including R$^3$Det[18], ReDet[14], S$^2$ANet[19], Faster-RCNN-O[6], Oriented RCNN[15], and RetinaNet-O[8]. For convenience, we test Faster-RCNN-O[6] and RetinaNet-

| Methods | backbone | PL | BD | BR | GTF | SV | LV | SH | TC | BC | ST | SBF | RA | HA | SP | HC | mAp |
|---|---|---|---|---|---|---|---|---|---|---|---|---|---|---|---|---|---|
| *Anchor-base(two-stage)* | | | | | | | | | | | | | | | | | |
| Roi-Trans.*[13] | R101 | 88.64 | 78.52 | 43.44 | 75.92 | 68.81 | 73.68 | 83.59 | 90.74 | 77.27 | 81.46 | 58.39 | 53.54 | 62.83 | 58.93 | 47.67 | 69.56 |
| CenterMap*[16] | R101 | 89.83 | 84.41 | 54.60 | 70.25 | 77.66 | 78.32 | 87.19 | 90.66 | 84.89 | 85.27 | 56.46 | 69.23 | 74.13 | 71.56 | 66.06 | 76.03 |
| SCRDet++*[17] | R101 | 90.05 | 84.39 | 55.44 | 73.99 | 77.54 | 71.11 | 86.05 | 90.67 | 87.32 | 87.08 | 69.62 | 68.90 | 73.74 | 71.29 | 65.08 | 76.81 |
| ReDet[14] | ReR50 | 88.79 | 82.64 | 53.97 | 74.00 | 78.13 | 84.06 | 88.04 | 90.89 | 87.78 | 85.75 | 61.76 | 60.39 | 75.96 | 68.07 | 63.59 | 76.25 |
| ReDet‡[14] | ReR50 | 88.81 | 82.48 | 60.83 | 80.82 | 78.34 | 86.06 | 88.31 | 90.87 | 88.77 | 87.03 | 68.65 | 66.90 | 79.26 | 79.71 | 74.67 | 80.10 |
| *Anchor-base(one-stage)* | | | | | | | | | | | | | | | | | |
| R$^3$Det*[18] | R152 | 89.80 | 83.77 | 48.11 | 66.77 | 78.76 | 83.27 | 87.84 | 90.82 | 85.38 | 85.51 | 65.67 | 62.68 | 67.53 | 78.56 | 72.62 | 76.47 |
| CSL*[20] | R152 | 90.13 | 84.43 | 54.57 | 68.13 | 77.32 | 72.98 | 85.94 | 90.74 | 85.95 | 86.36 | 63.42 | 65.82 | 74.06 | 73.67 | 70.08 | 76.24 |
| S$^2$A-Net[19] | R50 | 89.11 | 82.84 | 48.37 | 71.11 | 78.11 | 78.39 | 87.25 | 90.83 | 84.90 | 85.64 | 60.36 | 62.60 | 65.26 | 69.13 | 57.94 | 74.12 |
| S$^2$A-Net*[19] | R50 | 88.89 | 83.60 | 57.74 | 81.95 | 79.94 | 83.19 | 89.11 | 90.78 | 84.87 | 87.81 | 70.30 | 68.25 | 78.30 | 77.01 | 69.58 | 79.42 |
| *Anchor-free(one-stage)* | | | | | | | | | | | | | | | | | |
| BBAVectors*[27] | R101 | 88.63 | 84.06 | 52.13 | 69.56 | 78.26 | 80.40 | 88.06 | 90.87 | 87.23 | 86.39 | 56.11 | 65.62 | 67.10 | 72.08 | 63.96 | 75.36 |
| PIoU[26] | DLA-34 | 80.90 | 69.70 | 24.10 | 60.20 | 38.30 | 64.40 | 64.80 | 90.90 | 77.20 | 70.40 | 46.50 | 37.10 | 57.10 | 61.90 | 64.00 | 60.50 |
| DRN*[32] | H104 | 89.45 | 83.16 | 48.98 | 62.24 | 70.63 | 74.25 | 83.99 | 90.73 | 84.60 | 85.35 | 55.76 | 60.79 | 71.56 | 68.82 | 63.92 | 72.95 |
| CFA[31] | R101 | 89.26 | 81.72 | 51.81 | 67.17 | 79.99 | 78.25 | 84.46 | 90.77 | 83.40 | 85.54 | 54.86 | 67.75 | 73.04 | 70.24 | 64.96 | 75.05 |
| ProbIoU[25] | R50 | 89.09 | 72.15 | 46.92 | 62.22 | 75.78 | 74.70 | 86.62 | 89.59 | 78.35 | 83.15 | 55.83 | 64.01 | 65.50 | 65.46 | 46.23 | 70.04 |
| PolarDet[28] | R50 | 89.73 | 87.05 | 45.30 | 63.32 | 78.44 | 76.65 | 87.13 | 90.79 | 80.58 | 85.89 | 60.97 | 67.94 | 68.20 | 74.63 | 68.67 | 75.02 |
| PolarDet*[28] | R101 | 89.65 | 87.07 | 48.14 | 70.97 | 78.53 | 80.34 | 87.45 | 90.76 | 85.63 | 86.87 | 61.64 | 70.32 | 71.92 | 73.09 | 67.15 | 76.64 |
| FCOSR-S(ours) | Mobile v2 | 89.09 | 80.58 | 44.04 | 73.33 | 79.07 | 76.54 | 87.28 | 90.88 | 84.89 | 85.37 | 55.95 | 64.56 | 66.92 | 76.96 | 55.32 | 74.05 |
| FCOSR-S*(ours) | Mobile v2 | 88.60 | 84.13 | 46.85 | 78.22 | 79.51 | 77.00 | 87.74 | 90.85 | 86.84 | 86.71 | 64.51 | 68.17 | 67.87 | 72.08 | 62.52 | 76.11 |
| FCOSR-M(ours) | X50-32×4 | 88.88 | 82.68 | 50.10 | 71.34 | 81.09 | 77.40 | 88.32 | 90.80 | 86.03 | 85.23 | 61.32 | 68.07 | 75.19 | 80.37 | 70.48 | 77.15 |
| FCOSR-M‡(ours) | X50-32×4 | 89.06 | 84.93 | 52.81 | 76.32 | 81.54 | 81.81 | 88.27 | 90.86 | 85.20 | 87.58 | 68.63 | 70.38 | 75.95 | 79.73 | 75.67 | 79.25 |
| FCOSR-L(ours) | X101-64×4 | 89.50 | 84.42 | 52.58 | 71.81 | 80.49 | 77.72 | 88.23 | 90.84 | 84.23 | 86.48 | 61.21 | 67.77 | 76.34 | 74.39 | 74.86 | 77.39 |
| FCOSR-L‡(ours) | X101-64×4 | 88.78 | 85.38 | 54.29 | 76.81 | 81.52 | 82.76 | 88.38 | 90.80 | 86.61 | 87.25 | 67.58 | 67.03 | 76.86 | 73.22 | 74.68 | 78.80 |

Table 4. **Results on DOTA1.0 OBB Task.** H104 means Hourglass 104. * indicates multi-scale training and testing. ‡ means rotated test mode during multi-scale testing. The results with red and blue colors indicate the best and 2nd-best results of each column, respectively.

| Methods | backbone | PL | BD | BR | GTF | SV | LV | SH | TC | BC | ST | SBF | RA | HA | SP | HC | CC | mAp |
|---|---|---|---|---|---|---|---|---|---|---|---|---|---|---|---|---|---|---|
| RetinaNet-O♦[8] | R50 | 71.43 | 77.64 | 42.12 | 64.65 | 44.53 | 56.79 | 73.31 | 90.84 | 76.02 | 59.96 | 46.95 | 69.24 | 59.65 | 64.52 | 48.06 | 0.83 | 59.16 |
| FR-O●[6] | R50 | 71.89 | 74.47 | 44.45 | 59.87 | 51.28 | 68.98 | 79.37 | 90.78 | 77.38 | 67.50 | 47.75 | 69.72 | 61.22 | 65.28 | 60.47 | 1.54 | 62.00 |
| Mask R-CNN●[7] | R50 | 76.84 | 73.51 | 49.90 | 57.80 | 51.31 | 71.34 | 79.75 | 90.46 | 74.21 | 66.07 | 46.21 | 70.61 | 63.07 | 64.46 | 57.81 | 9.42 | 62.67 |
| DAFNe*[33] | R101 | 80.69 | 86.38 | 52.14 | 62.88 | 67.03 | 76.71 | 88.99 | 90.84 | 77.29 | 83.41 | 51.74 | 74.60 | 75.98 | 75.78 | 72.46 | 34.84 | 71.99 |
| FCOS♦[9] | R50 | 78.67 | 72.50 | 44.31 | 59.57 | 56.25 | 64.03 | 78.06 | 89.40 | 71.45 | 73.32 | 49.51 | 66.47 | 55.76 | 63.26 | 44.76 | 9.44 | 61.05 |
| ReDet●[14] | ReR50 | 79.20 | 82.81 | 51.92 | 71.41 | 52.38 | 75.73 | 80.92 | 90.83 | 75.81 | 68.64 | 49.29 | 72.03 | 73.36 | 70.55 | 63.33 | 11.53 | 66.86 |
| ReDet●⁑[14] | ReR50 | 88.51 | 86.45 | 61.23 | 81.20 | 67.60 | 83.65 | 90.00 | 90.86 | 84.30 | 75.33 | 71.49 | 72.06 | 78.32 | 74.73 | 76.10 | 46.98 | 76.80 |
| FCOSR-S(ours) | Mobile v2 | 80.05 | 76.98 | 44.49 | 74.17 | 51.09 | 74.07 | 80.60 | 90.87 | 78.40 | 75.01 | 53.38 | 69.35 | 66.33 | 74.43 | 59.22 | 13.50 | 66.37 |
| FCOSR-S*(ours) | Mobile v2 | 87.84 | 84.60 | 53.35 | 75.67 | 65.79 | 80.71 | 89.30 | 90.89 | 84.18 | 84.23 | 63.53 | 73.07 | 73.29 | 76.15 | 72.64 | 14.72 | 73.12 |
| FCOSR-M(ours) | X50-32×4 | 80.48 | 81.90 | 50.02 | 72.32 | 56.82 | 76.37 | 81.06 | 90.86 | 78.62 | 77.32 | 53.63 | 66.92 | 73.78 | 74.20 | 69.80 | 15.73 | 68.74 |
| FCOSR-M⁑(ours) | X50-32×4 | 80.85 | 83.89 | 53.36 | 76.24 | 66.85 | 82.54 | 89.61 | 90.87 | 80.11 | 84.27 | 61.72 | 72.90 | 76.23 | 75.28 | 70.01 | 35.87 | 73.79 |
| FCOSR-L(ours) | X101-64×4 | 80.58 | 85.25 | 51.05 | 70.83 | 57.77 | 76.72 | 81.09 | 90.87 | 78.07 | 77.60 | 51.91 | 68.72 | 75.87 | 72.61 | 69.30 | 31.06 | 69.96 |
| FCOSR-L⁑(ours) | X101-64×4 | 87.12 | 83.90 | 53.41 | 70.99 | 66.79 | 82.84 | 89.66 | 90.85 | 81.84 | 84.52 | 67.78 | 74.52 | 77.25 | 74.97 | 75.31 | 44.81 | 75.41 |

Table 5. **Results on DOTA1.5 OBB task.** The results of Faster R-CNN OBB (FR-O)[6], RetinaNet OBB (RetinaNet-O)[8], and Mask R-CNN[7] are reference from ReDet[14]. FCOS[9] result is from RotationDetection (https://github.com/yangxue0827/RotationDetection). ♦ and ● means one-stage and two-stage anchor-base methods, respectively. * indicates multi-scale training and testing. ⁑ means rotation test mode during multi-scale testing.

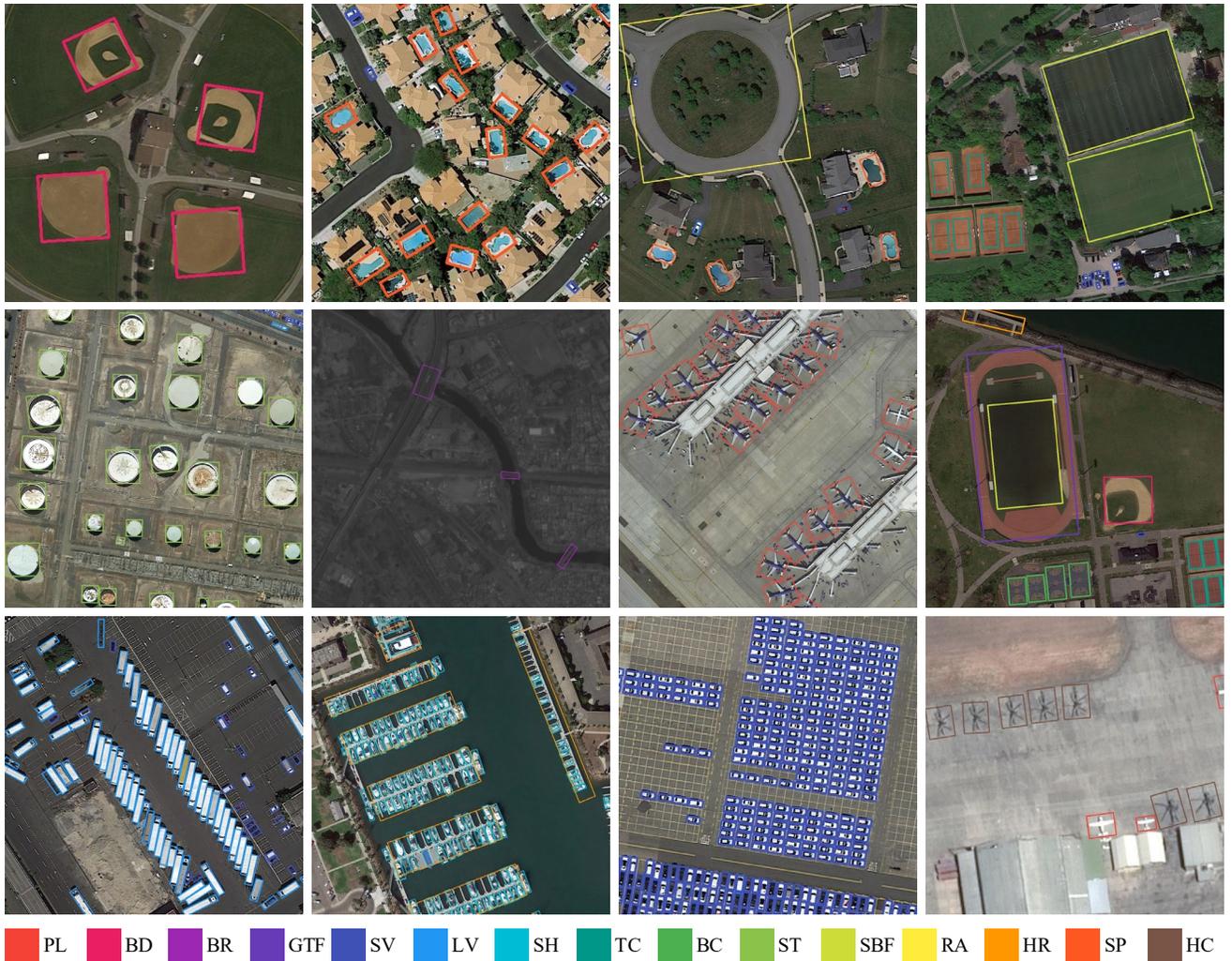

Figure 6. **The FCOSR-M detection result on DOTA1.0 test set.** The confidence threshold is set to 0.3 when visualizing these results.

O[8] models in the Oriented-RCNN[15] repository (https://github.com/jbwang1997/OBBDetection). The test result is shown in Figure 5. FCOSR-M exceeds almost the same speed S²ANet[19] and Oriented RCNN[15] **3.03** mAP and **1.28** mAP, respectively. FCOSR-S even achieves **74.05** mAP at a speed of **23.7** FPS, it become the fastest model currently. The FCOSR series models surpass the existing mainstream models in speed and accuracy, which also proves that through reasonable label assignment, even a very simple model can achieve excellent performance.

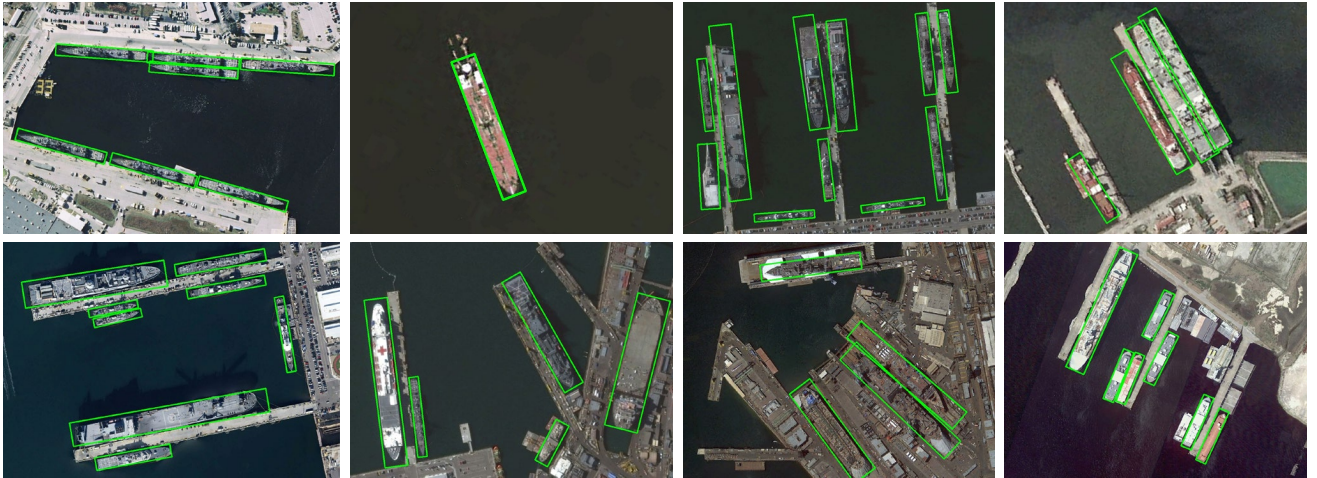

Figure 7. **The FCOSR-L detection result on HRSC2016.** The confidence threshold is set to 0.3 when visualizing these results.

| Methods | Backbone | mAP (07) | mAP (12) |
|---|---|---|---|
| PIoU[26] | DLA-34 | 89.20 | - |
| S$^2$A-Net[19] | R101 | 90.17 | 95.01 |
| ProbIOU[25] | R50 | 87.09 | - |
| DRN[32] | H104 | - | 92.70 |
| CenterMap[16] | R50 | - | 92.80 |
| BBAVectors[27] | R101 | 88.60 | - |
| PolarDet[28] | R50 | 90.13 | - |
| FCOSR-S(ours) | Mobile v2 | 90.08 | 92.67 |
| FCOSR-M(ours) | X50-32×4 | 90.15 | 94.84 |
| FCOSR-L(ours) | X101-64×4 | 90.14 | 95.74 |

Table 6. **Results on HRSC2016.** mAP 07 and 12 indicate VOC2007 metrics and VOC2012 metrics, respectively.

### 4.6. Comparison with state-of-the-arts

**Results on DOTA1.0.** As shown in Table 4, we compare FCOSR series with other state-of-the-arts methods on DOTA1.0 OBB task. FCOSR-L achieve **77.39** mAP, exceeding all single-scale methods and most multi-scale methods. Multi-scale FCOSR-M achieves **79.25** mAP, and the gap with S$^2$A-Net[19] was reduced to 0.17. Although there is still a certain gap compared with the anchor-base model, our algorithm achieves state-of-the-arts in anchor-free methods. We visualize a part of DOTA1.0 test set result in Figure 6.

**Results on DOTA1.5.** As shown in Table 5, we also conduct all experiments on the FCOSR series. However, there are currently few methods for evaluating the DOTA1.5 dataset, so we directly use some results in ReDet[14]. From a single-scale perspective, FCOSR-M and L models achieves **68.74** and **69.96** mAP, respectively. FCOSR-L outperform all single-scale models. From a multi-scale perspective, FCOSR-L achieves **75.41** mAP, which is the second best model on the DOTA1.5 test set.

**Results on HRSC2016.** As shown in Table 6, FCOSR series models surpass all anchor-free models, and achieve **95.74** mAP under the VOC2012 metrics. FCOSR series exceed 90 mAP under the VOC2007 metrics. FCOSR-L even surpasses S$^2$A-Net[19], which further proves that the anchor-free method we proposed already has the performance not weaker than the anchor-base method. The visualization of detection result is shown in Figure 7.

## 5. Conclusion

This paper proposed a one-stage anchor-free oriented detector, which consists of three parts: ellipse center sampling, fuzzy sample label assignment, and multi-level sampling. Ellipse center sampling provides a more suitable sampling area for rotated objects. The fuzzy sample label assignment method divides the sampling area of overlapping targets more reasonably. Multi-level sampling method alleviates the insufficient sampling problem of targets with large aspect ratios. Thanks to the simple architecture, FCOSR does not have any special computing unit for inferencing. Therefore, it is a very fast and easy deployment model to most platforms. Our experiments on light weight backbone have also shown satisfactory results. Extensive experiments on DOTA and HRSC2016 datasets demonstrate the effectiveness of our method.

**Acknowledgments**

This work was supported in part by the Key Scientific Technological Innovation Research Project by Ministry of Education; the National Natural Science Foundation of China under Grant 62171347，61877066, 61771379，62001355，62101405; the Foundation for Innovative Research Groups of the National Natural Science Foundation of China under Grant 61621005; the Key Research and Development Program in Shaanxi Province of China under Grant 2019ZDLGY03-05 and 2021ZDLGY02-08; the Science and Technology Program in Xi'an of China under Grant XA2020-RGZNTJ-0021; 111 Project.